\documentclass[letterpaper]{article}

\usepackage{aaai19}
\usepackage{times}
\usepackage{helvet}
\usepackage{courier}
\usepackage{url}
\usepackage{graphicx}
\usepackage{color}
\usepackage{amsmath}
\usepackage{amssymb}
\usepackage{mathtools}
\usepackage{enumitem}
\usepackage{scrextend}
\usepackage{array}
\usepackage{tabularx}
\usepackage{siunitx}

\usepackage{xcolor}
\usepackage{colortbl}

\usepackage{graphicx}

\setlist[enumerate]{label*=\arabic*.}

\usepackage{amsthm,xpatch}
\xpatchcmd{\@thm}{.}{:}{}{}
\theoremstyle{definition}
\newtheorem{definition}{Definition}
\newtheorem{theorem}{Theorem}
\newtheorem{example}{Example}

\newcommand{\mockalph}[1]{}

\title{ABox Abduction via Forgetting in $\mathcal{ALC}$ (Long Version)}
\author{Warren Del-Pinto\ and Renate A. Schmidt\\
School of Computer Science, The University of Manchester\\
Oxford Road, Manchester M13 9PL, United Kingdom
}

\nocopyright

\begin{document}
\maketitle

\begin{abstract}{
Abductive reasoning generates explanatory hypotheses for new observations using prior knowledge. This paper investigates the use of forgetting, also known as uniform interpolation, to perform ABox abduction in description logic ($\mathcal{ALC}$) ontologies. Non-abducibles are specified by a forgetting signature which can contain concept, but not role, symbols. The resulting hypotheses are semantically minimal and each consist of a set of disjuncts. These  disjuncts are each independent explanations, and are not redundant with respect to the background ontology or the other disjuncts, representing a form of hypothesis space. The observations and hypotheses handled by the method can contain both atomic or complex $\mathcal{ALC}$ concepts, excluding role assertions, and are not restricted to Horn clauses. Two approaches to redundancy elimination are explored for practical use: full and approximate. Using a prototype implementation, experiments were performed over a corpus of real world ontologies to investigate the practicality of both approaches across several settings.
}
\end{abstract}

\section{Introduction}
\noindent The aim of abductive reasoning is to generate explanatory hypotheses for new observations, enabling the discovery of new knowledge. 
Abduction was identified as a form of reasoning by C.S. Peirce \cite{PeirceHypothesis}. It has also become a recurring topic of interest in the field of AI, leading to work such as abductive extensions of Prolog for natural language interpretation \cite{StickelProlog,HobbsStickel}, the integration of abduction with induction in machine learning \cite{MooneyAbductionInduction} including work in the fields of inductive \cite{MuggletonBryant2000} and abductive logic programming \cite{KakasALP,OliverXHAIL} and statistical relational AI \cite{RaghavanBALPs}.

This paper focuses on abduction in description logic (DL) ontologies. These consist of a TBox of information about general entities known as \textit{concepts} and \textit{roles} and an ABox of assertions over instances of these concepts known as \textit{individuals}. DL ontologies are widely used to express background knowledge and as alternative data models for knowledge management. They are commonly used in fields such as AI, computational linguistics, bio-informatics and robotics. The need for abductive reasoning in ontologies was highlighted by \cite{Elsenbroich}. Use cases include hypothesis generation, diagnostics and belief expansion for which most current reasoning methods for ontologies are not suitable. 
This has led to a variety of work on abduction in DLs, including studies in $\mathcal{EL}$ \cite{BienvenuEL} and  applications such as ontology repair \cite{LambrixALC,KleinerDragisicLambrixTBox2014} and query explanation \cite{CalvaneseQueries}. For ABox abduction, methods in more expressive logics such as $\mathcal{ALC}$ and its extensions have been proposed  \cite{KlarmanABox2011,HallandBritzABox2012,Pukancova2017}. Similarly, a variety of work exists on TBox abduction \cite{KleinerDragisicLambrixTBox2014,HallandBritzKlarmanTBox2014}. However, few implementations and evaluations are available for abductive reasoning. Exceptions include the ABox abduction method of \cite{Du2014} in Datalog rewritable ontologies and a TBox abduction method using justification patterns \cite{Du2017}.

The aim of this paper is to investigate the use of \emph{forgetting} for ABox abduction in DL ontologies. Forgetting is a non-standard reasoning method that restricts ontologies to a specified set of symbols, retaining all entailments preservable in the restricted signature. It is also referred to as \emph{uniform interpolation} or \emph{second-order quantifier elimination}, and has been proposed as a method for abduction in different contexts \cite{DohertyLukasSzalas2001,SOQEBookApplications,WernhardAbduction2013,KoopmannNonStandardReasoning}. 
However, so far the forgetting-based approach has been insufficiently studied or applied, particularly in terms of preferred characteristics of abductive hypotheses and in the setting of large DL ontologies.

This work investigates the hypotheses obtained using forgetting-based abductive reasoning. These hypotheses are weakest sufficient conditions \cite{Lin2001}, related to the DL literature notion of semantic minimality \cite{HallandBritzKlarmanTBox2014}, meaning that they make the fewest assumptions necessary to explain an observation given the background knowledge. However, without additional steps, these hypotheses are not guaranteed to be consistent and are likely to be mostly redundant when the forgetting based approach is applied to large ontologies. In this work, additional constraints are investigated to capture these redundancies and practical methods for their removal are presented.

The main contributions of this paper are: (1) Forgetting-based ABox abduction in DL ontologies is explored and formalised. The aim is to compute hypotheses that do not contain unnecessary assumptions nor misleading, i.e. redundant, explanations. The need to eliminate redundancies from uniform interpolants is motivated and solved. (2) A practical method for this task is presented for $\mathcal{ALC}$. It computes hypotheses containing only abducible symbols. Non-abducibles are specified by a forgetting signature consisting of any set of concept, but not role, symbols. Both the observations and hypotheses may contain any atomic or complex $\mathcal{ALC}$ (or $\mathcal{ALC}\mu$) concepts, but cannot contain role assertions. An efficient annotation-based filtering method is proposed to eliminate redundancies from uniform interpolants. The method uses the forgetting tool LETHE which is shown to be applicable to ABox abduction, thereby answering an open question in \cite{KoopmannNonStandardReasoning}. However, the general framework could use any forgetting method designed for $\mathcal{ALC}$. (3) The method is evaluated empirically over a corpus of real-world ontologies. An approximate and a full approach to redundancy elimination are compared.

Proofs and additional examples can be found in the appendix.

\section{Problem Definition}\label{ProblemDefinitionSection}
Concepts in the description logic $\mathcal{ALC}$ have the following forms:
$\top \, \vert \, A \, \vert \, \lnot C \, \vert \, C \sqcup D \, \vert \, C \sqcap D \, \vert \, \forall r.C \, \vert \, \exists r.D$, where $A$ denotes a concept name, $C$ and $D$ are arbitrary $\mathcal{ALC}$ concepts and $r$ is a role name. \emph{Atomic} concepts are concept names, while concepts such as $\forall r.(A \sqcap B)$ are said to be \emph{complex}. A knowledge base or ontology $\mathcal{O}$ in $\mathcal{ALC}$ consists of a TBox and an ABox. The TBox consists of a set of general concept inclusions of the form $C \sqsubseteq D$, where $C$ and $D$ are any $\mathcal{ALC}$ concept. The ABox contains axioms $C(a)$ and  role assertions of the form $r(a,b)$, where $C$ is any $\mathcal{ALC}$ concept and $a$ and $b$ are individuals. The signature of $X$, denoted as $sig(X)$, is the set of all concept and role names occurring \mbox{in $X$}, where $X$ can be any ontology or axiom.

The aim of abduction is to compute a hypothesis to explain a new observation. This paper focuses on the following form of the ABox abduction problem.

\begin{definition}\label{AbductionProblem}
\textbf{Abduction in Ontologies.} \textit{Let $\mathcal{O}$ be an ontology and $\psi$ a set of ABox axioms, where $\psi$ does not contain role assertions, such that $\mathcal{O} \not\models {\perp}$, $\mathcal{O}, \psi \not\models {\perp}$ and $\mathcal{O} \not\models \psi$. Let $\mathcal{S}_A$ be a set of symbols called \emph{abducibles} which contains all role symbols in $(\mathcal{O}, \psi)$. The \emph{abduction problem} is to find a hypothesis $\mathcal{H}$ as a disjunction of ABox axioms, without role assertions, that contains only those symbols specified in $\mathcal{S}_A$ such that: \emph{(i)} $\mathcal{O}, \mathcal{H} \not\models {\perp}$, \emph{(ii)} $\mathcal{O}, \mathcal{H} \models \psi$, \emph{(iii)} $\mathcal{H}$ does not contain \emph{inter-disjunct} redundancy i.e., there is no disjunct $\alpha_i$ in $\mathcal{H}$ such that $\mathcal{O}, \alpha_i \models \alpha_1 \sqcup ... \sqcup \alpha_{i-1} \sqcup \alpha_{i+1} \sqcup ... \sqcup \alpha_n$ and \emph{(iv)} for any $\mathcal{H}'$ satisfying conditions (i)--(iii) where $sig(\mathcal{H}') \subseteq \mathcal{S}_A$, if $\mathcal{O, H} \models \mathcal{O, H}'$ then $\mathcal{O, H}' \models \mathcal{O, H}$.}
\end{definition}

The set of abducibles $\mathcal{S}_A$ defines the subset of symbols in the ontology that may appear in the hypothesis $\mathcal{H}$. \mbox{Here, $\mathcal{S}_A$} must contain all role symbols in $(\mathcal{O}, \psi)$ and both the observation $\psi$ and $\mathcal{H}$ may not contain role assertions. For our approach, the language of $\mathcal{ALC}$ must be extended to include disjunctive ABox assertions over multiple individuals, and in some specific cases fixpoints \cite{CalvaneseFixpoints1999} to represent cyclic results. These will be discussed alongside the proposed method.

The rationale for the problem conditions is to focus efforts on computing informative hypotheses. Otherwise, the search space for hypotheses would be too large. Defining the set of abducibles $\mathcal{S}_A$ allows a user to focus on explanations containing specific information represented as symbols, utilising their own knowledge of the problem domain.

Conditions (i) and (ii) of Definition 1 are standard requirements in most abductive reasoning tasks. \mbox{Condition (i)} requires that all generated hypotheses $\mathcal{H}$ are consistent with the background knowledge in the ontology $\mathcal{O}$. Otherwise ${\perp}$ would be entailed from which everything follows. \mbox{Condition (ii)} ensures that $\mathcal{H}$ explains the observation $\psi$ when added to the background knowledge in $\mathcal{O}$. 

Conditions (iii) and (iv) capture two distinct notions. Condition (iii) ensures that each of the disjuncts in the hypothesis $\mathcal{H}$ are independent explanations \cite{Konolige1992} for the observation $\psi$. That is, there are no disjuncts in $\mathcal{H}$ that express information that is the same or more specific than that which is already expressed by the other disjuncts in $\mathcal{H}$. This also excludes disjuncts that are simply inconsistent with the background knowledge as a special case, since if for a disjunct $\alpha \in \mathcal{H}$ we have $\mathcal{O}, \alpha \models \perp$ then everything follows. Condition (iii) is referred to as \emph{inter-disjunct redundancy}. The example below illustrates its use:

\begin{example}\label{DisjunctRedundancyExample}
Let $\mathcal{O} = \{B \sqsubseteq D, A \sqsubseteq C \sqcup B, C \sqsubseteq D, A \sqsubseteq D, E \sqsubseteq \, \perp \}$, $\psi = D(a)$ and $\mathcal{S}_A = \{A, B, C, E\}$. Consider the hypotheses: $\mathcal{H}_1 = (B \sqcup C)(a)$ and $\mathcal{H}_2 = (A \sqcup B \sqcup C \sqcup E)(a)$. Both satisfy conditions (i), (ii) \mbox{and (iv)}. However, $\mathcal{H}_2$ contains two redundant disjuncts: $A(a)$ and $E(a)$. $A(a)$ provides no new information over the other disjuncts: $\mathcal{O}, A(a) \models (B \sqcup C)(a)$, while $E(a)$ is inconsistent with $\mathcal{O}$. While not stronger than $\mathcal{H}_1$, $\mathcal{H}_2$ is unnecessarily complex. A user may have the false impression that $E(a)$ is a valid explanation for $\psi$, or that $A(a)$ is an independent avenue of explanation compared to $(B \sqcup C)(a)$. \mbox{Condition (iii)} excludes these redundancies: for $\mathcal{H}_2$ it is the case that $\mathcal{O}, A(a) \models (B \sqcup C)(a)$ and also $\mathcal{O}, E(a) \models \perp$ and thus everything follows. As a result, $\mathcal{H}_2$ is excluded and $\mathcal{H}_1$ is returned as the solution.
\end{example}

As condition (iii) requires that each disjunct be consistent with the ontology $\mathcal{O}$, condition (i) is not strictly needed: $\mathcal{O, H} \not\models \perp$ follows if condition (iii) is satisfied. However, as consistency is a key condition in most abduction contexts it is useful to emphasise it as a separate characteristic.

Condition (iv) captures the notion of \emph{semantic minimality} \cite{HallandBritzKlarmanTBox2014} under the background knowledge $\mathcal{O}$. It restricts hypotheses to those that make the fewest assumptions necessary to explain the observation $\psi$ given $\mathcal{O}$. This is shown in the example below. 

\begin{example}\label{StrongSemanticMinimalityExample}
Let $\mathcal{O} = \{A \! \sqsubseteq \! B, B \! \sqsubseteq \! C\}$, $\psi = \{C(a)\}$ and $\mathcal{S}_A = \{A, B\}$. Consider the hypotheses $\mathcal{H}_1 = B(a)$ and $\mathcal{H}_2 = A(a)$. Both satisfy the conditions in Definition 1(i) and (ii). However, hypothesis $\mathcal{H}_2$ does not satisfy (iv), since $\mathcal{O, H}_2 \models \mathcal{O, H}_1$, but the reverse does not hold. Thus, $\mathcal{H}_2$ is a stronger or ``less minimal" hypothesis than $\mathcal{H}_1$.
\end{example}

From this, it can be seen that condition (iv) rejects semantically stronger hypotheses. It should be noted that, unlike some other settings such as \cite{HallandBritzKlarmanTBox2014}, here $\mathcal{H}$ can contain disjunctions. Thus, redundant disjuncts must be considered separately, as in \mbox{condition (iii)}, since condition (iv) does not account for these. 

With these conditions, the aim of this work is to compute a semantically minimal hypothesis consisting of all disjuncts that each represent an independent explanation of the observation $\psi$, none of which overlaps with either the background knowledge or the other disjuncts. 

Definition 1 does not remove all choices between or redundancies in the forms taken by each disjunct in $\mathcal{H}$ if they are equivalent under $\mathcal{O}$. For example, condition (iv) does not account for conjunctively joined redundancies that follow directly from $\mathcal{O}$. If Example 2 is extended so that the \mbox{axiom $C \sqsubseteq D$} is in $\mathcal{O}$ and the signature of abducibles $\mathcal{S}_A$ also contains $D$, then $\mathcal{H}_3 = (B \sqcap D)(a)$ is also a valid hypothesis under conditions (i), (ii) and (iv). While $\mathcal{H}_3$ is not stronger than $\mathcal{H}_1$, it contains a form of redundancy: $D(a)$. 

To eliminate these redundancies and simplify the disjuncts themselves may require the use of preference criteria over the disjuncts in $\mathcal{H}$. As there are a variety of methods for defining and realising preference handling \cite{MayerPirri1996,PinoPreferenceRels,DelgrandePreferenceSurvey} we do not discuss this aspect. Here, the focus is on computing the space of independent explanations, rather than ensuring each takes the simplest form. 

\section{Forgetting and Uniform Interpolation}\label{ForgettingSection}
\noindent Forgetting is a process of finding a compact representation of an ontology by hiding or removing subsets of symbols within it. Here, the term \textit{symbols} refers to concept and role names present in the ontology.  The symbols to be hidden are specified in the \emph{forgetting signature} $\mathcal{F}$, which is a subset of symbols in the ontology $\mathcal{O}$. The symbols in $\mathcal{F}$ should be removed from $\mathcal{O}$, while preserving all entailments of $\mathcal{O}$ that can be represented using the signature $sig(\mathcal{O})$ without $\mathcal{F}$. The result is a new ontology, which is a uniform interpolant:

\begin{definition}\label{UniformInterpolationInALC}
\textbf{Uniform Interpolation in $\mathcal{ALC}$ \cite{LutzWolter2011}.} \textit{Let $\mathcal{O}$ be an $\mathcal{ALC}$ ontology and $\mathcal{F}$ a set of symbols to be forgotten from $\mathcal{O}$. Let $\mathcal{S}_A = sig(\mathcal{O}) \setminus \mathcal{F}$ be the complement of $\mathcal{F}$. The \emph{uniform interpolation problem} is the task of finding an ontology $\mathcal{V}$ such that the following conditions hold: \textnormal{(i)} $sig(\mathcal{V}) \subseteq \mathcal{S}_A$, \textnormal{(ii)} for any axiom $\beta$: $\mathcal{O} \models \beta$ iff $\mathcal{V} \models \beta$ provided that $sig(\beta) \subseteq \mathcal{S}_A$. The ontology $\mathcal{V}$ is a \emph{uniform interpolant} of $\mathcal{O}$ for the signature $\mathcal{S}_A$. We also say that $\mathcal{V}$ is the \emph{result of forgetting} $\mathcal{F}$ from $\mathcal{O}$.}
\end{definition}

Uniform interpolants are strongest necessary entailments, in general, it holds that:

\begin{theorem}\label{UniformInterpolants}
$\mathcal{V}$ is a uniform interpolant of ontology $\mathcal{O}$ for $\mathcal{S}_A$ iff $\mathcal{V}$ is a strongest necessary entailment of $\mathcal{O}$ in $\mathcal{S}_A$. 
\end{theorem}

This means that for any ontology $\mathcal{V}'$, if $sig(\mathcal{V}') \subseteq \mathcal{S}_A$ and $\mathcal{V}'\models\mathcal{V}$, then $\mathcal{V}\models\mathcal{V}'$. Of the methods for uniform interpolation in $\mathcal{ALC}$, e.g., \cite{LudwigKonev2014,KoopmannABoxes}, our abduction method uses the resolution-based method developed by Koopmann and Schmidt [\citeyear{KoopmannFixpoints,KoopmannABoxes,KoopmannNonStandardReasoning}]. 

Here, this method is referred to as $Int_{\mathcal{ALC}}$. Motivations for utilising $Int_{\mathcal{ALC}}$ include the fact that it can perform forgetting for $\mathcal{ALC}$ with ABoxes \cite{KoopmannABoxes}, making it suitable for the setting in this paper. Furthermore, in theory the result of forgetting (and abduction) can involve an infinite chain of axioms. Using $Int_{\mathcal{ALC}}$, such chains can be finitely represented using fixpoint operators. In practice, these are rarely required: in previous work only 7.2\% of uniform interpolants contained cycles \cite{KoopmannFixpoints}. $Int_{\mathcal{ALC}}$ can also handle disjunctive ABox assertions which are not representable in pure $\mathcal{ALC}$. These will be needed for some abduction cases involving multiple individuals. In terms of efficiency, the size of the forgetting result is constrained to at most a double exponential bound with respect to the input ontology and $Int_{\mathcal{ALC}}$ is guaranteed to terminate \cite{KoopmannABoxes}.

The method $Int_{\mathcal{ALC}}$ has two properties that are also essential to the proposed abduction method. (i) \emph{Soundness}: any ontology $\mathcal{O}'$ returned by applying $Int_{\mathcal{ALC}}$ to an ontology $\mathcal{O}$ is a uniform interpolant. (ii) \emph{Interpolation Completeness}: if there exists a uniform interpolant $\mathcal{O}'$ of ontology $\mathcal{O}$, then the result of $Int_{\mathcal{ALC}}$ is an ontology $\mathcal{V}$ such that $\mathcal{V} \equiv \mathcal{O}'$. Thus, for any $\mathcal{ALC}$ ontology $\mathcal{O}$ and any forgetting signature $\mathcal{F}$, $Int_{\mathcal{ALC}}$ always returns a finite uniform interpolant.
\begin{figure}[h!]
\noindent \hspace{-2pt}\fbox{
\small
	\parbox{\linewidth-1.1em}{
		{
		
		 \textbf{Resolution:} \hfill \underline{$C_1 \lor A(t_1) \hspace{20pt} C_2 \lor \lnot A(t_2)$}
		
		\hspace{16.5em} $(C_1 \lor C_2)(\sigma)$ \\
		
		\textbf{Role Propagation:} \hfill \underline{$C_1 \lor (\forall r.D_1)(t_1) \hspace{20pt} C_2 \lor Qr.D_2(t_2)$}
		
		\hspace{12em}$(C_1 \lor C_2)\sigma \lor Qr.D_{12}(t_1\sigma)$ \\

		\textbf{$\exists$-Role Restriction Elimination:} \hfill \underline{$C \lor (\exists r.D)(t) \hspace{20pt} \lnot D(x)$}
		
		\hspace{19em} $C$ \\
		
				\textbf{Role Instantiation:} \hfill \underline{$C_1 \lor (\forall r.D)(t_1) \hspace{20pt} r(t_2, b)$}
		
		\hspace{17em} $C_1\sigma \lor D(b)$
		
}

$D_1$ and $D_2$ are definer symbols,  Q $\in \{\forall, \exists\}$, $\sigma$ is the unifier of $t_1$ and $t_2$ if it exists, $D_{12}$ is a new definer symbol for $D_1 \sqcap D_2$ and no clause contains more than one negative definer literal of the form $\lnot D(x)$, and none of the form $\lnot D(a)$.
	}
}
\caption{$Int_{\mathcal{ALC}}$ rules utilised in our abduction method.}
\end{figure}

The method $Int_{\mathcal{ALC}}$ relies on the transformation of the ontology to a normal form given by a set of clauses of concept literals. The inference rules of the forgetting calculus utilised in $Int_{\mathcal{ALC}}$ are shown in Figure 1. Definer symbols are introduced to represent concepts that fall under the scope of a quantifier. Resolution inferences are restricted to concepts in $\mathcal{F}$ or definer symbols. Once all possible inferences have been made, any clauses containing symbols in $\mathcal{F}$ are removed and the definer symbols are eliminated resulting in an ontology $\mathcal{O}'$ that is free of all symbols in $\mathcal{F}$. A discussion of this calculus and the associated method, including proofs, can be found in \cite{KoopmannABoxes}.

We will also need the following notions. Each premise in an application of an inference rule in $Int_{\mathcal{ALC}}$ is referred to as a \emph{parent} of the conclusion of the rule. The \emph{ancestor} relation is defined as the reflexive, transitive closure of the parent relation. For example, the premises $\{A\sqsubseteq B, C \sqsubseteq A, \lnot B(a)\}$ are expressed as the clauses: $\{\lnot A(x) \lor B(x), \lnot C(x) \lor A(x), \lnot B(a)\}$. For a forgetting signature $\mathcal{F} = \{B, A\}$, resolution between $\lnot A(x) \lor B(x)$ and $\lnot B(a)$ gives $\lnot A(a)$. Resolution between $\lnot A(a)$ and $\lnot C(x) \lor A(x)$ gives $\lnot C(a)$. The axioms $A\sqsubseteq B$ and $\lnot B(a)$ are the \emph{parents} of the axiom $\lnot A(a)$ and the \emph{ancestors} of $\lnot C(a)$.

In this paper, we focus on ABox abduction where the set of abducibles includes all role symbols. Non-abducibles are specified by the \emph{forgetting signature} $\mathcal{F}$ which contains only concept symbols occurring in the ontology $\mathcal{O}$ or observation $\psi$. The proposed method utilises $Int_{\mathcal{ALC}}$ to compute semantically minimal hypotheses via forgetting and contrapositive reasoning, exploiting: $\mathcal{O}, \mathcal{H} \models \psi$ iff $\mathcal{O}, \lnot \psi \models \lnot \mathcal{H}$ where $\mathcal{O}$ is an ontology and $\psi$, $\mathcal{H}$ are (ABox) axioms.

\section{A Forgetting-Based Abduction Method}\label{MethodSection}
The abduction algorithm takes as \textbf{input} an $\mathcal{ALC}$ ontology $\mathcal{O}$, an observation $\psi$ as a set of ABox axioms and a forgetting signature $\mathcal{F}$. 

Several assumptions are made regarding this input. The method $Int_{\mathcal{ALC}}$ does not cater for negated role assertions as can be seen in Figure 1, and the form of role forgetting in $Int_{\mathcal{ALC}}$ is not complete for abduction. As a result, $\psi$ cannot contain role assertions and $\mathcal{F}$ is restricted to concept symbols in $sig(\mathcal{O} \cup \psi)$. Correspondingly, the signature of abducibles $\mathcal{S}_A$ must contain all role symbols occurring in $sig(\mathcal{O} \cup \psi)$. Also, if $\mathcal{F}$ does not contain at least one symbol in the observation $\psi$, the semantically minimal hypothesis will simply be $\psi$ itself, i.e., $\mathcal{H} = \psi$. This is reflected in the fact that no inferences would occur between $\mathcal{O}$ and $\lnot\psi$ under $Int_{\mathcal{ALC}}$. To avoid this trivial hypothesis, $\mathcal{F}$ should contain at least one concept symbol in the signature of $\psi$. In the event that $\mathcal{F}$ contains concepts that occur within a cycle in $\mathcal{O}$, the forgetting result obtained using $Int_{\mathcal{ALC}}$ may contain greatest fixpoints \cite{KoopmannFixpoints} to finitely represent infinite forgetting solutions. For our method, this means that the abduction result may contain least fixpoints due to the negation of greatest fixpoints under contraposition. In these cases, the output language would be $\mathcal{ALC}\mu$.

The \textbf{output} is a hypothesis $\mathcal{H} = \alpha_1(a_1) \sqcup ... \sqcup \alpha_n(a_n)$ containing only the abducible symbols $\mathcal{S}_A = sig(\mathcal{O} \cup \psi) \setminus \mathcal{F}$, that satisfies the conditions (i)--(iv) in Definition \ref{AbductionProblem}. Note that $\mathcal{H}$ may be a disjunctive assertion over several individuals, again motivating the need to extend $\mathcal{ALC}$ with these. 

The algorithm reduces the task of computing abductive hypotheses for the observation $\psi$ to the task of forgetting, using the following steps:

\begin{enumerate}[label=\textbf{(\arabic*)},leftmargin=7mm]
\item Compute the uniform interpolant $\mathcal{V} = \{\beta_1,..., \beta_n\}$ of $(\mathcal{O},\lnot\psi)$ with respect to the forgetting signature $\mathcal{F}$. 

\item Extract the set $\mathcal{V}^* \subseteq \mathcal{V}$ by omitting axioms $\beta_i \in \mathcal{V}$ such that $\mathcal{O}, \beta_1,...,\beta_{i-1},\beta_{i+1},...,\beta_n \models \beta_i$.

\item Obtain the hypothesis $\mathcal{H}$ by negating the set $\mathcal{V}^*$.
\end{enumerate}

In more detail, the input observation $\psi$ takes the form of a set of ABox axioms: $\psi = \{C_1(a_1), ... , C_k(a_k)\}$ where the $C_i$ are $\mathcal{ALC}$ concepts and the $a_i$ are individuals. The negation takes the form \mbox{$\lnot\psi = \lnot C_1(a_1) \sqcup ... \sqcup \lnot C_k(a_k)$}. The forgetting method $Int_{\mathcal{ALC}}$ is used to compute the uniform interpolant $\mathcal{V}$ of $(\mathcal{O}, \lnot\psi)$ by forgetting the concept names in $\mathcal{F}$, i.e., $\mathcal{V} = (\mathcal{O},\lnot\psi)^{-\mathcal{F}}$.

If forgetting was used in isolation, the negation of $\mathcal{V}$ would be the hypothesis for $\psi$ under contraposition. However, this is only guaranteed to satisfy conditions (ii) \mbox{and (iv)} of Definition 1: since $\mathcal{V}$ is the strongest necessary entailment of $(\mathcal{O}, \lnot\psi)$ in $\mathcal{S}_A$ as in Theorem \ref{UniformInterpolants}, its negation would be the weakest sufficient condition \cite{Lin2001,DohertyLukasSzalas2001}. Thus the hypothesis would be semantically minimal in $\mathcal{S}_A$, but would not necessarily satisfy condition (i), consistency, nor condition (iii), absence of inter-disjunct redundancy. In practice most of the disjuncts will be redundant, as the experimental results show (Table 2). In the case that there is no suitable hypothesis, an inconsistent or ``false" hypothesis will be returned since all of the axioms in $\mathcal{V}$ would follow directly from $\mathcal{O}$.

Step (2) therefore omits information in $\mathcal{V}$ that follows from the background knowledge $\mathcal{O}$ together with other axioms in $\mathcal{V}$ itself. This check is the dual of Definition 1(iii), and therefore eliminates inter-disjunct redundancies such as those in Example 1. The result is a \emph{reduced uniform interpolant} $\mathcal{V}^*$ which takes the form \mbox{$\mathcal{V}^* = \{\beta_1(a_1), ... , \beta_k(a_k)\}$} where each $\beta_i$ is an $\mathcal{ALC}(\mu)$ concept. 


If an axiom $\beta_i$ is redundant, it is removed from $\mathcal{V}$ immediately. For the following disjuncts, the check is performed against the remaining axioms in $\mathcal{V}$. This avoids discarding too many axioms: if multiple axioms express the same information, i.e. are equivalent under $\mathcal{O}$, one of them should be retained in the final hypothesis $\mathcal{H}$. For example, if two axioms $\beta_i$ and $\beta_j$ are equivalent under $\mathcal{O}$, but are otherwise not redundant, only one of them is discarded. The order in which the axioms are checked can be random, or can be based on some preference relation \cite{MayerPirri1996}.

In Step (3) the reduced uniform interpolant $\mathcal{V}^*$ is negated, resulting in the hypothesis $\mathcal{H}$. Thus, each disjunct $\alpha_i$ in $\mathcal{H}$ is the negation of an axiom $\beta_i$ in $\mathcal{V}^*$, i.e., \mbox{$\alpha_i \equiv \lnot\beta_i$}.

The soundness and completeness of the method are made explicit in the following theorem.

\begin{theorem}\label{SoundnessAndCompleteness}
Let $\mathcal{O}$ be an $\mathcal{ALC}$ ontology, $\psi$ an observation as a set of ABox axioms, excluding role assertions, and $\mathcal{S}_A$ a set of abducible symbols that includes all role symbols in $\mathcal{O}, \psi$ and $\mathcal{S}_A \subseteq sig(O, \psi)$. \emph{(i) \textbf{Soundness}}: The hypothesis $\mathcal{H}$ returned by the method is a disjunction of ABox axioms such that sig($\mathcal{H}) \subseteq \mathcal{S}_A$ and $\mathcal{H}$ satisfies Definition 1(i)-(iv). \mbox{\emph{(ii) \textbf{Completeness}}}: If there exists a hypothesis $\mathcal{H}'$ such that $sig(\mathcal{H}') \subseteq \mathcal{S}_A$ and $\mathcal{H}'$ satisfies Definition 1(i)--(iv), then the method returns a hypothesis $\mathcal{H}$ such that $\mathcal{O, H} \equiv \mathcal{O, H}'$.
\end{theorem}

\begin{theorem}\label{WorstComplexity}
In the worst case, computing a hypothesis $\mathcal{H}$ using our method has 3EXPTIME upper bound complexity for running time and the size of $\mathcal{H}$ can be double exponential in the size of $(\mathcal{O}, \psi)$.
\end{theorem}

\section{Practical Realisation}\label{FilteringSection}
For redundancy elimination, Step (2) requires checking the relation $\mathcal{O}, \mathcal{V}\setminus \beta_i \not\models \beta_i$ for every axiom $\beta_i$ in $\mathcal{V}$. Since entailment checking in $\mathcal{ALC}$ has exponential complexity and $\mathcal{V}$ is in the worst case double exponential in the size of $(\mathcal{O}, \lnot\psi)$, this step has a 3EXPTIME upper bound which is very expensive particularly for large ontologies. Regardless, Step (2) is essential; without it there will be a large number of inter-disjunct redundancies (Definition 1(iii)) in the hypotheses obtained.
 This is reflected in the experiments (Table 2).


To obtain a computationally feasible implementation of Step (2), the number of entailment checks performed must be reduced. Our implementation of this step begins by tracing the dependency of axioms in $\mathcal{V}$ on the negated observation $\lnot\psi$. An axiom $\beta$ is defined as \emph{dependent} upon $\lnot\psi$ if in the derivation using $Int_{\mathcal{ALC}}$ it has at least one \emph{ancestor} axiom in $\lnot\psi$. The set of axioms dependent on $\lnot\psi$ is in general a superset of the reduced uniform interpolant $\mathcal{V}^*$ and is referred to as $\mathcal{V}^*_{app}$, i.e., an \emph{approximation} of $\mathcal{V}^*$.

In this paper, dependency tracing is achieved by using \emph{annotations}, similar to \cite{KazakovAnnotations,KoopmannSOQE,PenalozaLeanKernels}. These take the form of fresh concept names that do not occur in the signature of the ontology nor the observation. Annotations act as labels that are disjunctively appended to existing axioms. They are then used to trace which axioms are the ancestors of inferred axioms. This relies on the fact that the annotation concept is not included in the forgetting signature $\mathcal{F}$. Thus, it will carry over from the parent to the result of any inference in $Int_{\mathcal{ALC}}$, as formalised in the following property:

\begin{theorem}\label{AnnotationCarries}
Let $\mathcal{O}$ be an ontology, $\psi$ an observation as a set of ABox axioms, $\mathcal{F}$ a forgetting signature and $\mathcal{\ell}$ an annotator concept added as an extra disjunct to each clause in the clausal form of $\lnot\psi$ where $\mathcal{\ell} \not\in sig(\mathcal{O \cup \psi})$ and $\mathcal{\ell} \not\in \mathcal{F}$. For every axiom  $\beta$ in the uniform interpolant $\mathcal{V} = (\mathcal{O}, \lnot\psi)^{-\mathcal{F}}$, $\beta$ is dependent on $\lnot\psi$ iff $\mathcal{\ell} \in sig(\beta)$.
\end{theorem}

Therefore, the presence of the annotation concept in the signature of an inferred axiom indicates that the axiom has at least one ancestor in $\lnot\psi$. Since the aim is to trace dependency specifically on $\lnot\psi$, only clauses that are part of $\lnot\psi$ need to be annotated. As it is not important which specific clauses in $\lnot\psi$ were used in the derivation of dependent axioms, only one annotation concept name is required. This will be referred to as $\mathcal{\ell}$. Using this technique, the process of extracting $\mathcal{V}^*_{app}$ from the uniform interpolant $\mathcal{V}$ is a matter of removing all axioms in $\mathcal{V}$ that do not contain $\mathcal{\ell}$. Then, $\mathcal{\ell}$ can be replaced with ${\perp}$ to obtain the annotation-free set $\mathcal{V}^*_{app}$.


Since this annotation based filtering is sound, i.e., it only removes axioms that are not dependent on $\psi$, as these are directly derivable from $\mathcal{O}$ and are thus guaranteed to be redundant, it can be used at the start of Step (2) to \mbox{compute $\mathcal{V}^*_{app}$}. To guarantee the computation of the reduced uniform interpolant $\mathcal{V}^*$, the entailment check in Step (2) must then be performed for each axiom $\beta \in \mathcal{V}^*_{app}$ to eliminate any redundancies not captured by the annotation-based filtering. Since some axioms may have multiple derivations, they can contain the annotation concept but still be redundant with respect to Definition 1. For example:

\begin{example}\label{ExistingIndividualExtraStep}
Let $\mathcal{O} = \{A \sqsubseteq C, B \sqsubseteq C, A \sqcap D \sqsubseteq {\perp}, D(a)\}$ and $\psi = C(a)$. The annotated form of $\lnot\psi$ is \mbox{$\lnot\psi = \mathcal{\ell} \sqcup \lnot C(a)$}. Using $\mathcal{F} = \{C\}$, the result of Step (1) is $\mathcal{V} = \{A \sqcap D \sqsubseteq {\perp}, D(a), (\mathcal{\ell} \sqcup \lnot A)(a), (\mathcal{\ell} \sqcup \lnot B)(a) \}$. Note: no inference is made with $D(a)$, since $\mathcal{D} \not\in \mathcal{F}$. In Step (2) extracting all axioms with annotations and setting $\mathcal{\ell} = {\perp}$ gives the set $\{\lnot A(a), \lnot B(a)\}$. Despite $\lnot A(a)$ being derivable using $\lnot\psi$, it follows from the original ontology $\mathcal{O}$ and is therefore redundant with respect to Definition 1(iii). This can now be removed via the entailment check in Step (2).
\end{example}

This method of filtering out redundancies has several advantages. First, it is not specific to $\mathcal{ALC}$ and can be applied if the abduction method is later extended to more expressive logics. Second, by removing axioms that are not dependent on $\psi$, the method reduces the cost of Step (2) since checking the signature of each axiom for the presence of $\mathcal{\ell}$ is linear in the size of $\mathcal{V}$. In the worst case $\mathcal{V}^*_{app}$ could be equal to $\mathcal{V}$ and a double exponential number of entailment checks would still be required.
In practice, this is unlikely as $\mathcal{V}^*_{app}$ is usually a small fraction of $\mathcal{V}$ as shown by the experiments (Table 2). In these cases, each redundancy eliminated from $\mathcal{V}$ to $\mathcal{V}^*_{app}$ replaces an exponential check with a linear one.

The entailment checks that must be performed on $\mathcal{V}^*_{app}$ to compute $\mathcal{V}^*$ may still be costly in the event that many axioms are dependent on $\psi$ in $\mathcal{V}$. Therefore, we propose that in some cases it may be pragmatic to relax the allowed hypotheses by negating $\mathcal{V}^*_{app}$ instead of the reduced uniform interpolant $\mathcal{V}^*$ itself. In this case, an additional check, $\mathcal{O, H} \not\models \perp$, is required to rule out inconsistent hypotheses if all of the axioms in $\mathcal{V}^*_{app}$ are redundant. This approximate approach results in a hypothesis $\mathcal{H}_{app}$ which satisfies conditions (i), (ii) and (iv) in Definition 1, but not condition (iii). The results in Table 2 illustrate the effect in practice.

To summarise, we suggest two realisations of Step (2) of the proposed abduction method: (a) \emph{approximate} filtering, which computes an approximation of the hypothesis $\mathcal{H}_{app}$ by negating $\mathcal{V}^*_{app}$, (b) \emph{full} filtering, which performs the entailment check in Step (2) for each axiom in $\mathcal{V}^*_{app}$ to \mbox{obtain $\mathcal{V}^*$} and thus $\mathcal{H}$ which is guaranteed to fully satisfy Definition 1. Note that for setting (b), the approximation step is still used to reduce the overall cost of Step (2).

\section{Experimental Evaluation}\label{Experiments}

\begin{table}[ht]
\small
\centering 
\begin{tabular*}{\linewidth}{| l@{\extracolsep{\fill}} | r | r | r | r | r |}
\hline
Ontology \hspace{1pt} & DL & TBox & ABox & Num. & Num. \hspace{-1pt} \\
Name &    & Size & Size & Concepts & Roles \\
\hline
BFO & $\mathcal{EL}$ & 52 & 0 & 35 & 0 \\
LUBM & $\mathcal{EL}$ & 87 & 0 & 44 & 24 \\
HOM & $\mathcal{EL}$ & 83 & 0 & 66 & 0 \\
DOID & $\mathcal{EL}$ & 7892 & 0 & 11663 & 15 \\
SYN & $\mathcal{EL}$ & 15352 & 0 & 14462 & 0 \\
ICF & $\mathcal{ALC}$ & 1910 & 6597 & 1597 & 41 \\
Semintec & $\mathcal{ALC}$ & 199 & 65189 & 61 & 16 \\ 
OBI & $\mathcal{ALC}$ & 28888 & 196 & 3691 & 67 \\
NATPRO & $\mathcal{ALC}$ & 68565 & 42763 & 9464 & 12 \\
\hline
\end{tabular*}
\caption{Characteristics of the experimental corpus.}
\end{table}

\begin{table*}[ht]
\small
\centering 
\noindent\begin{tabularx}{\textwidth}{@{}c@{\extracolsep{\fill}}c c c c c c c c c c c@{}}
\hline
Ont. & \multicolumn{3}{c}{{Mean Time Taken /s}} & \multicolumn{3}{c}{{Max Time Taken /s}}& \multicolumn{2}{c}{{Mean Redund. Removed}} & \multicolumn{2}{c}{ Size $\mathcal{H}$ /disjuncts} & Mean \% of \\
Name &  $\mathcal{V}^*_{app}$ & $\mathcal{V}^*$ & $\mathcal{V}^*$ no app. & $\mathcal{V}^*_{app}$ & $\mathcal{V}^*$ & $\mathcal{V}^*$ no app. & $\mathcal{V}\rightarrow\mathcal{V}^*_{app}$ &  $\mathcal{V}^*_{app}\rightarrow\mathcal{V}^*$ & Mean & Max & $\mathcal{H}_{app}$ Redund. \\
\hline
BFO & 0.01 & 0.01 & 0.09 & 0.01 & 0.07 & 0.14 & 52 & 0 & 1.97 & 4 & 0 \\
LUBM & 0.02 & 0.03 & 0.30 & 0.11 & 0.16 & 1.21 & 90 & 0.80 & 2.73 & 11 & 29.30 \\
HOM & 0.03 & 0.05 & 0.18  & 0.40 & 0.54 & 0.86 & 82 & 0.03 & 2.07 & 13 & 1.45 \\
DOID & 0.44 & 1.09 & 1071.35 & 1.11 & 6.98 & 1095.07 & 7891 & 0 & 7.23 & 104 & 0 \\
SYN & 0.95 & 3.92 & 2421.96 & 2.33 & 61.52 & 2593.13 & 15351 & 0.03 & 20.63 & 457 & 0.15 \\
ICF & 0.30 & 0.56 & t.o. & 0.52 & 1.58 & t.o. & 8505 & 0 & 2.30 & 7 & 0 \\
Semin. & 3.13 & 5.12 & t.o. & 9.29 & 15.36 & t.o. & 72827 & 0.03 & 3.60 & 10 & 0.83\\
OBI* & 3.82 & 32.17 & t.o. & 25.18 & 95.37 & t.o. & 29191 & 6.48 & 52.48 & 161 & 12.35 \\
NATP. & 26.54 & 179.70 & t.o. & 39.51 & 544.50 & t.o. & 111318 & 0.03 & 48.70 & 204 & 0.06 \\
\hline

\end{tabularx}
\caption{Results for 30 observations using a forgetting signature size of 1. * indicates that LETHE did not terminate within the 300s time limit in at least one case, ``t.o." indicates that the experiment was terminated after several days runtime. The size of $\mathcal{H}$ reported is that obtained via full computation of $\mathcal{V}^*$. Times shown are the total times taken to return $\mathcal{H}$ (or $\mathcal{H}_{app}$).}
\end{table*}

A Java prototype was implemented using the OWL-API\footnote{http://owlapi.sourceforge.net/} and the forgetting tool LETHE which implements the $Int_{\mathcal{ALC}}$ method.\footnote{http://www.cs.man.ac.uk/~koopmanp/lethe/index.html}. Using this, two experiments were carried out over a corpus of real world ontologies, which were preprocessed into their $\mathcal{ALC}$ fragments. Axioms not representable in $\mathcal{ALC}$, such as number restrictions of the form $\leq nR.C$ where $R$ is a role symbol and $C$ is a concept symbol, were removed. Others were represented using appropriate $\mathcal{ALC}$ axioms where possible. For example, a range restriction $\exists$r$^{-}.\top \sqsubseteq C$ is converted to $\top \sqsubseteq \forall r.C$, where $r^{-}$ is the inverse role of $r$. The choice of ontologies was based on several factors. They must be consistent, parsable using LETHE and the OWL API and must vary in size to determine how this impacts performance. Since many real-world ontologies are encoded in less expressive DLs such as $\mathcal{EL}$, the corpus was also split between $\mathcal{EL}$ and $\mathcal{ALC}$ to determine if the performance over $\mathcal{EL}$ suffers as a result of the additional capabilities of the method for $\mathcal{ALC}$. The final corpus contains ontologies from the NCBO Bioportal and OBO repositories,\footnote{https://bioportal.bioontology.org/}\footnote{http://www.obofoundry.org/} and the LUBM \cite{LUBMBenchmark} and Semintec ontologies.\footnote{http://www.cs.put.poznan.pl/alawrynowicz/semintec.htm } The corpus is summarised in Table 1. The experiments were performed on a machine using a 4.00GHz Intel Core i7-6700K CPU and 16GB RAM. 

For each ontology, 30 consistent, non-entailed observations were randomly generated using any $\mathcal{ALC}$ concepts from the associated ontology, some of which were combined using $\mathcal{ALC}$ operators to encourage variety. The aim was to emulate the information that may be observed in practice for each ontology, while adhering to the requirements for $\psi$ expressed in Definition 1. As the current prototype uses the OWL-API, which does not allow disjunctive assertions over multiple individuals, the experiments here are limited to observations involving one individual. For the filtering in Step (2), the preference relation used in these experiments was simply based on order of appearance of each disjunct.

For the first experiment, $\mathcal{F}$ was set to one random concept symbol from $sig(\psi)$. The assumption was that users may first seek the most general hypothesis, i.e., the semantically minimal hypothesis for the largest set of abducibles. This allows the user to pursue stronger hypotheses subsequently by forgetting further symbols from the initial hypothesis. This experiment is therefore also representative of incremental abduction steps using a small $\mathcal{F}$. The second experiment was performed over the DOID, ICF and SYN ontologies to evaluate the performance as the size of $\mathcal{F}$ increases. These ontologies were used as they have a sufficiently large signature of concepts and LETHE did not time out when forgetting in any case. In all cases, at least one symbol from $\psi$ was present in $\mathcal{F}$ to avoid trivial hypotheses. 

In both experiments, the approaches based on (a) \emph{approximate} and (b) \emph{full} filtering were compared for the same observations and same random selection of $\mathcal{F}$. Thus, the tradeoff between the additional time for entailment checking and redundancy in the final hypothesis is evaluated. In all cases, LETHE was subject to a 300 second time limit.

Table 2 shows the results for the first experiment. For the smaller ontologies, the difference in time taken between the approximate and full filtering was small. For the larger ontologies the cost of the full filtering was more pronounced, taking 313\%, 742\% and 577\% longer across the SYN, OBI and NATPRO ontologies respectively. In all cases, it can be seen that the annotation-based filtering eliminated the majority of redundancies. In three cases (BFO, DOID, ICF), for all 30 observations the result of the approximation, $\mathcal{V}^*_{app}$, contained no redundancies and thus $\mathcal{H}_{app} = \mathcal{H}$. For the other ontologies, in most cases $\mathcal{V}^*_{app}$ contained few redundancies in both absolute terms and relative to the size of the final hypothesis. For the LUBM and OBI ontologies, however, these redundancies made up a more significant portion of $\mathcal{V}^*_{app}$.

The full filtering setting still uses the annotation-based method as a preprocessing step. To assess the benefit of this preprocessing, results for applying the entailment check in Step (2) directly to $\mathcal{V}$ instead of $\mathcal{V}^*_{app}$ were collected and are shown in the ``$\mathcal{V}^*$ no app." columns. For the largest $\mathcal{EL}$ ontologies, the time taken increased significantly e.g. taking 98,189\% longer for the DOID ontology. For all of the $\mathcal{ALC}$ ontologies the experiments were terminated after several days runtime, i.e., it took at least several hours to compute a single hypothesis on average.  This indicates that the annotation-based filtering significantly reduces the time taken, particularly over large or more expressive ontologies.

\begin{figure}[h!]
\includegraphics[width=8.4cm]{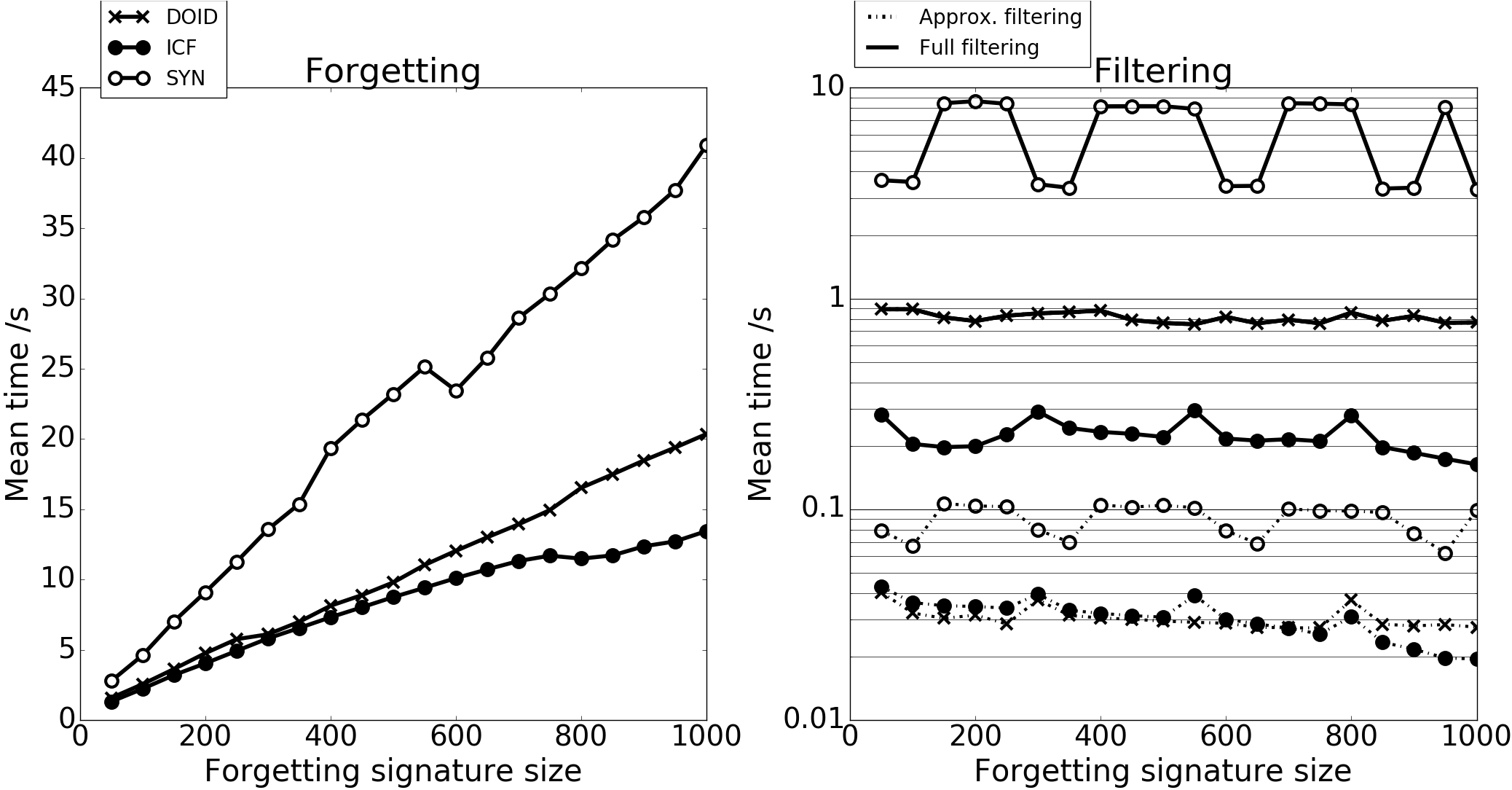}
\caption[time_graph]{Mean forgetting and filtering times with varying $\mathcal{F}$ signature sizes for the ICF, DOID and SYN ontologies.}
\label{sig_size}
\end{figure}

Figure \ref{sig_size} shows the results of the second experiment. The time taken for the forgetting step, Step (1), increased almost linearly with the size of $\mathcal{F}$. 
This was expected due to a higher number of inferences needed to compute $\mathcal{V}$. The time taken for filtering, Step (2), did not increase with the size of $\mathcal{F}$. However, for each ontology, maxima were observed for different sizes of $\mathcal{F}$. This implies that certain symbols increase the filtering time if they appear in $\mathcal{F}$. Forgetting commonly used concepts results in more inferences and a larger $\mathcal{V}$, which may explain the maxima as the annotation-based filtering depends solely on the number of axioms in $\mathcal{V}$. The size of $\mathcal{V}^*_{app}$ will also increase in these cases, leading to more exponential entailment checks for full filtering. The full filtering took an average of 27, 11 and 70 times longer than the approximate case for the DOID, ICF and SYN ontologies respectively. This indicates that the cost of the full  entailment check increased with the size of the ontology, particularly the size of the TBox, and not the size \mbox{of $\mathcal{F}$}.

In 100\% of cases for both experiments the hypotheses were represented without fixpoints, indicating that cyclic, semantically minimal hypotheses seem rare in practice.

\section{Discussion}
The use of forgetting for abduction has been suggested in classical logics \cite{DohertyLukasSzalas2001,SOQEBookApplications,WernhardAbduction2013}, and a form of TBox abduction \cite{KoopmannNonStandardReasoning}. Our work extends on these suggestions in several ways. As the focus is on large DL ontologies, and not small theories in classical logics, an interpretable hypothesis cannot be obtained by negating the forgetting \mbox{result $\mathcal{V}$} as most of it will be redundant (Table 2). Thus, we gain insight into the redundancies in $\mathcal{V}$ in terms of abductive notions, such as \cite{Konolige1992,HallandBritzKlarmanTBox2014}, resulting in Definition 1(iii) and (iv). Efficient redundancy removal is then achieved via the annotation-based filtering. The overall approach, including two options emphasising (a) practicality and (b) full redundancy removal, is then evaluated over a corpus of real-world ontologies. This is the first realisation and evaluation of a practical forgetting-based approach to ABox abduction in DL ontologies.

Restricting inferences in $Int_{\mathcal{ALC}}$ to axioms dependent on $\lnot\psi$, rather than filtering the output, was considered. However, this would not circumvent the need to perform entailment checking, as illustrated in Example \ref{ExistingIndividualExtraStep}. Second, computing the full uniform interpolant $\mathcal{V}$ has an interesting use case: iterative abduction. For example:

\begin{example}\label{ExampleIterativeAbduction}
Let $\mathcal{O} = \{A \sqsubseteq C, B \sqsubseteq C, C \sqsubseteq D\}$ and $\psi = D(a^*)$. In Step (1), using $\mathcal{F} $$ = $$ \{D\}$ results in $\mathcal{V} $$ = $$ \{A \sqsubseteq C, B \sqsubseteq C, \lnot C(a^*)\}$. Steps (2)--(3) result in $\mathcal{H} = C(a^*)$. Now,  forgetting $\mathcal{F}_2 = \{C\}$ from $\mathcal{V}$ of the previous iteration results in $\mathcal{V}_2 = \{\lnot A(a^*), \lnot B(a^*)\}$. Repeating Steps (2) and (3) gives $\mathcal{H}_2 = (A \sqcup B)(a^*)$, which is stronger than $\mathcal{H}$ and is the same as the result of computing the uniform interpolant of $(\mathcal{O}, \lnot\psi)$ using $\mathcal{F} = \{D,C\}$, but will be more efficient. 
\end{example}

This iterative process enables hypothesis refinement, and has potential synergy with induction. Data could inform the selection of new forgetting signatures to find stronger hypotheses following from prior likely hypotheses: a cycle of abduction, deduction and induction. 

Limitations include the lack of role assertions in the observations and hypotheses, due to the inability of $Int_{\mathcal{ALC}}$ to handle negated role assertions, and the incompleteness of role forgetting for abduction, as illustrated by the following:

\begin{example}\label{ExampleRoleRestriction}
Let $\mathcal{O} = \{C \sqsubseteq \exists r.D\}$ and $\psi = \exists r.D(a)$. Using $\mathcal{F} = \{r\}$ the result of Step (1) is $\mathcal{V} = \emptyset$. This is due to the fact that no inferences are possible on the symbol $D$, since resolution is restricted to $\mathcal{F}$. Thus, the hypothesis obtained is $\mathcal{H} = \emptyset$, while the expected result is $\mathcal{H} = C(a)$.
\end{example}

With the use of nominals, this limitation can be overcome. Options include the use of other forgetting approaches \cite{ZhaoForgetting2015,ZhaoForgetting2016} or the extension of $Int_{\mathcal{ALC}}$.

It should be noted that methods such as \cite{KlarmanABox2011,Pukancova2017} can already handle role assertions. The former is a purely theoretical work, which restricts the abductive observations and solutions to $\mathcal{ALE}$: the fragment of $\mathcal{ALC}$ without disjunctions of concepts and allowing only atomic negation. The method of \cite{Pukancova2017} performs abductive reasoning up to $\mathcal{ALCHO}$, restricting observations and hypotheses to atomic and negated atomic concept and role assertions. This method considers syntactic, but not semantic, minimality, though the authors note the importance of semantic minimality in practical applications.


\section{Conclusion and Future Work}
In this paper, a practical method for ABox abduction in $\mathcal{ALC}$ ontologies was presented. The method computes semantically minimal hypotheses with independent disjuncts to explain observations, where both may contain complex $\mathcal{ALC}$ concepts but not role assertions, and the set of abducibles must contain all role symbols. The practicality of the method, including the proposed annotation-based filtering, was evaluated over a corpus of real-world ontologies. To the best of our knowledge, this is the first method that computes such hypotheses efficiently in large ontologies. The ability to produce a semantically minimal space of independent explanations will likely be beneficial in real-world applications. For example, this can provide engineers with multiple, non-redundant suggestions for fixing errors in an ontology or explaining negative query results, even over large knowledge bases. For scientific investigation using ontologies, the ability to produce independent avenues of explanation starting with the least assumptions necessary captures the essence of scientific hypothesis formation. The ability to refine these hypotheses via repeated forgetting also provides a goal-oriented, potentially data driven, way to derive stronger insights from the hypotheses produced.

Future work will include removing the restriction on role assertions. Also, though forgetting in DLs can be applied to a form of TBox abduction \cite{KoopmannNonStandardReasoning}, the hypotheses take the \mbox{form $\top \sqsubseteq \alpha_1 \sqcup ... \sqcup \alpha_n$} where each $\alpha$ is an $\mathcal{ALC}$ concept. Thus, the problem of determining inter-disjunct redundancy  and the proposed approach differ in several aspects. This will be investigated, as will the iterative abduction use case.

{\small
\bibliography{AAAIReferences}
}
\bibliographystyle{aaai}

\section{Appendix}
\subsection{Examples}
For the inferences shown in the following examples, the rules in $Int_{\mathcal{ALC}}$ are referred to as follows, where X and Y are arbitrary axioms denoted by a number:

(1) res(X, Y): resolution

(2) role\_ prop(X, Y): role propagation

(3) exis\_ elim(X, Y): existential role elimination

Here is an example demonstrating the full procedure for an ABox observation, including the normal form used by $Int_{\mathcal{ALC}}$ and all inferences:

\begin{example}\label{ExampleUIFiltering}
Consider the ontology $\mathcal{O}$ containing the axioms:

\vspace{5pt}

\emph{Pogona $\sqsubseteq \exists$livesIn.(Woodland $\sqcap$ Arid)}

\emph{Sloth $\sqsubseteq$ Mammal}

\emph{Woodland $\sqsubseteq$ Habitat}

\emph{PineWoods $\sqsubseteq$ Woodland}

\vspace{5pt}

\noindent and an observation \emph{$\psi = \exists$livesIn.Woodland(Gary)}. Using a forgetting signature \emph{$\mathcal{F} = \{$Woodland$\}$}, the first two steps in computing a hypothesis are as follows. \mbox{Step (1)}: Negate $\psi$ and annotate it to obtain \emph{$\lnot\psi = (\mathcal{\ell} \sqcup \forall$livesIn.$\lnot$Woodland)(Gary)} and add this to $\mathcal{O}$. Convert $(\mathcal{O},\lnot\psi)$ to the normal form required by $Int_{\mathcal{ALC}}$. This results in the clause set:

\vspace{5pt}

  \emph{1. $(\lnot$Pogona $\sqcup \, \exists$livesIn.$D_1)(x)$}
  
  \emph{2. $(\lnot D_1 \, \sqcup $ Woodland$)(x)$}
  
  \emph{3. $(\lnot D_1 \, \sqcup$ Arid$)(x)$}
  
  \emph{4. $(\lnot$Sloth $\sqcup$ Mammal$)(x)$}
  
  \emph{5. $(\lnot$Woodland $\sqcup$ Habitat$)(x)$}
  
  \emph{6. $(\lnot$PineWoods $\sqcup$ Woodland$)(x)$}
  
  \emph{7. $(\mathcal{\ell} \sqcup \forall$livesIn.$D_2)$(Gary)}
  
  \emph{8. $(\lnot D_2 \sqcup \lnot$Woodland$)(x)$}

\vspace{5pt}

\noindent where $D_1$ and $D_2$ are definer symbols. Step (2): Apply $Int_{\mathcal{ALC}}$ to this clause set to compute the uniform \mbox{interpolant $\mathcal{V}$}. The inferences are as follows:  

\vspace{5pt}

\begin{tabularx}{\linewidth}{@{}l*{10}{l}}
9. \emph{$(\lnot D_1 \, \sqcup$ Habitat$)(x)$} & res(2,5) \\

10. \emph{$(\lnot D_2 \, \sqcup \lnot$PineWoods$)(x)$} & res(6,8) \\

11. \emph{$(\lnot$PineWoods $ \sqcup$ Habitat$)(x)$} & res(5,6) \\

12. \emph{$(\mathcal{\ell} \sqcup \lnot$Pogona $\sqcup \exists$livesIn.$D_{12})$(Gary)} & role\_ prop(1,7) \\

13. \emph{$(\lnot D_{12} \sqcup D_1)(x)$} \\

14. \emph{$(\lnot D_{12} \sqcup D_2)(x)$} \\

15. \emph{$(\lnot D_{12} \sqcup $ Woodland$)(x)$} & res(2,13) \\

16. \emph{$(\lnot D_{12} \sqcup \lnot$PineWoods$)(x)$} & res(10,14)\\

17. \emph{$(\lnot D_{12} \sqcup \lnot$Woodland$)(x)$} & res(8,14) \\

18. \emph{$\lnot D_{12}(x)$} & res(15,17) \\

19. \emph{$(\mathcal{\ell} \sqcup \lnot$Pogona)(Gary)} & exis\_ elim(12,18) \\
\end{tabularx}

\vspace{5pt}

\noindent Here, all inferences possible under $Int_{\mathcal{ALC}}$ have been made. Now, definers are eliminated by the reverse of the Ackermann rules \emph{\cite{KoopmannABoxes}}, following which all clauses containing symbols in $\mathcal{F} = \{$Woodland$\}$ or any remaining definer symbols are deleted. The resulting uniform interpolant is

\vspace{5pt}

$\mathcal{V} =$ \{ \emph{Pogona$ \sqsubseteq \exists$livesIn.(Habitat $\sqcap$ Arid)},

\emph{\hphantom{$\mathcal{V} =$ \{} Sloth $\sqsubseteq$Mammal},

\emph{\hphantom{$\mathcal{V} =$ \{} PineWoods $\sqsubseteq$ Habitat},

\emph{\hphantom{$\mathcal{V} =$ \{} $(\mathcal{\ell} \sqcup \lnot$Pogona)(Gary)},

\emph{\hphantom{$\mathcal{V} =$ \{} $(\mathcal{\ell} \sqcup \forall$livesIn.$\lnot$PineWoods)(Gary)} \}

\vspace{5pt}

\noindent Step (2): Remove all axioms with a signature that does not contain the annotation concept $\mathcal{\ell}$. The first, second and third axioms are all discarded and it is clear that these follow from the original \mbox{ontology $\mathcal{O}$}. The annotation concept can then be eliminated by setting $\mathcal{\ell} = {\perp}$, leaving the approximate reduced uniform interpolant:

\vspace{5pt}

\emph{$\mathcal{V}^*_{app} = \{\lnot$Pogona(Gary), $\forall$livesIn.$\lnot$PineWoods(Gary)$\}$}.

\vspace{5pt}

\noindent For the full filtering procedure, the remaining two axioms in $\mathcal{V}^*_{app}$ are then subject to the entailment check in Step (2): for each $\beta_i \in \mathcal{V}$, if $\mathcal{O}, \beta_1,..., \beta_{i-1}, \beta_{i+1},..., \beta_n \models \beta_i$ then $\beta_i$ is redundant and is removed from $\mathcal{V}$. In this case, neither of the two axioms is redundant since:

\vspace{5pt}

\emph{$\mathcal{O}, \lnot$Pogona(Gary)$\not\models \forall$livesIn.$\lnot$PineWoods(Gary)} and \\

\emph{$\mathcal{O}, \forall$livesIn.$\lnot$PineWoods(Gary)$\not\models \lnot$Pogona(Gary)}

\vspace{5pt}

\noindent Thus, $\mathcal{V}^*_{app} = \mathcal{V}^*$ and the result of Step (2) is:

\vspace{5pt}

\emph{$\mathcal{V}^* = \{\lnot$Pogona(Gary), $\forall$livesIn.$\lnot$PineWoods(Gary)$\}$}.

\vspace{5pt}

\noindent Step (3): negate the set $\mathcal{V}^*$. This results in the hypothesis: 

\vspace{5pt}

\emph{$\mathcal{H} = [$Pogona $\sqcup \exists$livesIn.PineWoods$]$(Gary)}\\

\vspace{5pt}


\end{example}

\subsubsection{Cyclic Example}
For the experimental corpus, no hypotheses containing fixpoints were observed. This means that no cyclic, semantically minimal hypotheses were obtained for each combination of observation and random signature of abducibles. This is consistent with the results obtained by \cite{KoopmannFixpoints}, given that here the cycles would need to occur specifically over axioms dependent \mbox{on $\psi$}. 

However, despite the rarity of these hypotheses it is still important to consider the meaning of such hypotheses and how they might occur. Below is a small example of such a case using the method described in this paper:

\begin{example}\label{ExampleCycles}
Consider the following ontology $\mathcal{O}$: 

\vspace{5pt}

\emph{Mammal} $\sqsubseteq \exists $\emph{hasParent.Mammal}

\vspace{5pt}

\noindent and an observation $\psi = \{\lnot$\emph{Mammal}$(a^*)\}$ with a forgetting signature \emph{$\mathcal{F} = \{$Mammal$\}$}. In Step (1), the observation is negated, annotated to obtain $\lnot\psi = (\mathcal{\ell} \, \sqcup \, $\emph{Mammal} $)(a^*)$, and is added to $\mathcal{O}$. $Int_{\mathcal{ALC}}$ is applied as follows:

\vspace{5pt}

\begin{tabularx}{\linewidth}{@{}l*{10}{l}}

1. \emph{$(\lnot$Mammal $\sqcup \, \exists$hasParent.$D_1)(x)$} \\
2. \emph{$(\lnot D_1 \sqcup$ Mammal$)(x)$} \\
3. \emph{$(\mathcal{\ell} \, \sqcup \, $Mammal$)(a^*)$}\\

4. \emph{$(\lnot D_1 \sqcup \exists$hasParent.$D_1)(x)$} & res(1,2)\\
5. \emph{$(\mathcal{\ell} \sqcup \exists$hasParent.$D_1)(a^*)$} & res(1,3) \\

\end{tabularx}

\vspace{5pt}

\noindent At this point, all inferences have been made. Now definer symbols are eliminated and clauses containing symbols in $\mathcal{F} = \{$Mammal$\}$ are removed. The elimination of $D_1$ results in the introduction of a greatest fixpoint operator, representing a potentially infinite chain under the \emph{hasParent} relation in axioms 4 and 5. The resulting uniform interpolant is 

\vspace{5pt}

$\mathcal{V} = $ \{ $(\mathcal{\ell} \sqcup \exists$hasParent.$\nu X.(\exists$hasParent$.X))(a^*)$

\vspace{5pt}

\noindent where $\nu X$ represents a greatest fixpoint. In Step (2), the reduced uniform interpolant is simply: 

\vspace{5pt}

$\mathcal{V}^* = (\mathcal{\ell} \sqcup \exists$hasParent.$\nu X.(\exists$hasParent$.X))(a^*)$

\vspace{5pt}

\noindent In Step (3), the annotation is discarded by setting $\mathcal{\ell} = {\perp}$ before negating $\mathcal{V}^*$ to obtain the following hypothesis:

\vspace{5pt}

$\mathcal{H} = \forall$hasParent.$\mu X.(\exists$hasParent.$X)(a^*)$

\vspace{5pt}

\noindent where $\mu X$ represents a least fixpoint. 

\end{example}

The need to introduce the fixpoint operator in this example can be seen by seen by comparing the elimination of non-cylic and cyclic definers under Ackermann's lemma \cite{KoopmannFixpoints}:

\vspace{5pt}

{\small
\begin{tabularx}{\linewidth}{@{}c*{10}{c}}

\textbf{Non-cyclic Definer Elimination} & \textbf{Cyclic Definer Elimination} \\

\underline{$\mathcal{T} \cup \{D \sqsubseteq C\}$} & \underline{$\mathcal{T} \cup \{D \sqsubseteq C\}$} \\

$\mathcal{T}^{D \mapsto C}$ & $\mathcal{T}^{D \mapsto \nu X.C[X]}$

\end{tabularx}
}

\noindent where $\mathcal{T}$ is a set of axioms, $C$ is an $\mathcal{ALC}$ concept, $D$ is a definer symbol and $\nu X$ represents a greatest fixpoint, where $X$ is concept variable. In the non-cyclic case, it is assumed that $D$ does not appear in $C$, while the opposite is true in the cyclic case. Thus, the introduction of a greatest fixpoint operator is due to the presence of axiom 4 in the above example.

To interpret the intuition and meaning behind the fixpoint hypothesis \mbox{$\mathcal{H} = \forall$hasParent.$\mu X.(\exists$hasParent.$X)(a^*)$} from the above example, consider the non-finite form of $\mathcal{H}$ without fixpoints:

\vspace{5pt}
\noindent $\mathcal{H} = (\forall$hasParent.$\perp \sqcup \, \forall$hasParent.($\forall$hasParent$.\!\perp) \sqcup ... )(a^*)$
\vspace{5pt}

\noindent Effectively this means that, if $a^*$ is not a Mammal as in $\psi$, then it must ``not have a parent" or "must have a parent who does not have a parent..." and so on. In this limited ontology, this is the semantically minimal hypothesis not involving the concept Mammal.

\subsection{Proofs of Theorems}
\subsubsection{Theorem 1}

First the following connection between uniform interpolants and strongest necessary conditions or entailments is proven: \\

\noindent {\textnormal{\textbf{Theorem 1:}} \textit{$\mathcal{V}$ is a uniform interpolant of ontology $\mathcal{O}$ for $\mathcal{S}_A$ iff $\mathcal{V}$ is the strongest necessary entailment of $\mathcal{O}$ in $\mathcal{S}_A$.}

\vspace{5pt}

\noindent \textnormal{\textbf{Proof:} $\mathcal{V} \models \mathcal{V}$ for any $\mathcal{V}$. Since $sig(\mathcal{V}) \subseteq \mathcal{S}_A$, then $\mathcal{O}\models\mathcal{V}$ by the reverse direction of Definition 2(ii). To show that $\mathcal{V}$ is the strongest necessary entailment of $\mathcal{O}$, let $\mathcal{V}'$ be any uniform interpolant of $\mathcal{O}$ such that $\mathcal{V}' \models \mathcal{V}$. Since $\mathcal{O} \models \mathcal{V}'$ and $sig(\mathcal{V}') \subseteq \mathcal{S}_A$, it follows by Definition 2(ii) that $\mathcal{V}\models\mathcal{V}'$. Thus, $\mathcal{V} \equiv \mathcal{V}'$ for any $\mathcal{V}'$. Now let $\mathcal{V}$ be the strongest necessary entailment of $\mathcal{O}$ in the signature $\mathcal{S}_A$. Trivially, this means that $\mathcal{V}$ satisfies condition (i). Let $\beta$ be an axiom such that $\mathcal{V} \models \beta$. Since $\mathcal{V}$ is a set of entailments of $\mathcal{O}$, it follows that if $\mathcal{V} \models \beta$ then $\mathcal{O} \models \beta$. As $\mathcal{V}$ is also the strongest set of entailments of $\mathcal{O}$, it follows that for any other axiom $\beta$ such that $\mathcal{O} \models \beta$ then $\mathcal{V} \models \beta$. Thus both directions of condition (ii) are satisfied and $\mathcal{V}$ is a uniform interpolant.} \\

\subsubsection{Soundness and Completeness}
Here, the soundness and completeness of the method with respect to the abduction problem outlined in Definition 1 are proved. Note, this is first proved for the general case without the annotation-based preprocessing step in Step (2), i.e., by applying the entailment check directly to $\mathcal{V}$ and not $\mathcal{V}^*_{app}$. Following this, the soundness of the annotation-based filtering is proved, which ensures that the overall soundness of the method is retained when the annotation-based preprocessing is used.

 Let $\mathcal{O}, \mathcal{V}, \mathcal{V}^*, \mathcal{S}_A$ and $\psi$ be defined as in Section 5. First,  Lemmas 1--5 cover key properties of the uniform interpolant $\mathcal{V}$ and the reduced uniform interpolant $\mathcal{V}^*$ which are useful in proving the soundness and completeness of the abduction method.\\

\noindent\textnormal{\textbf{Lemma 1:}} \textit{$\mathcal{O}, \lnot\psi \models \mathcal{V}$}

\vspace{5pt}

\noindent\textnormal{\textbf{Proof:}} \textnormal{The soundness of $Int_{\mathcal{ALC}}$ for computing uniform interpolants has been proven in \cite{KoopmannFixpoints,KoopmannABoxes}. Thus, the set $\mathcal{V}$ is a uniform interpolant of the input $(\mathcal{O},\lnot\psi)$ and satisfies the conditions in Definition \ref{UniformInterpolationInALC}. The property in Lemma 1 then follows from Theorem 1: if $\mathcal{V}$ is the strongest necessary entailment of $\mathcal{O}$ in a signature $\mathcal{S}_A$, then trivially $\mathcal{O}, \lnot\psi \models \mathcal{V}$.} \\

Lemmas 2, 3, 4 and 5 follow from the definition of Step (2) of the method: the reduction of the uniform interpolant to only the set $\mathcal{V}^*$ of axioms such that for each $\beta_i \in \mathcal{V}^*$, $\mathcal{O}, \beta_1,..., \beta_{i-1}, \beta_{i+1},..., \beta_n \not\models \beta_i$.\\

\noindent\textnormal{\textbf{Lemma 2:}} $\mathcal{O}, \lnot\psi \models \mathcal{V}^*$

\vspace{5pt}

\noindent\textnormal{\textbf{Proof:}} \textnormal{Given that $\mathcal{O}, \lnot\psi \models \mathcal{V}$ and $\mathcal{V}^* \subseteq \mathcal{V}$, it then follows that $\mathcal{O}, \lnot\psi \models \mathcal{V}^*$.}\\







\noindent \textbf{Lemma 3:} $\mathcal{O} \not\models \beta$ for every $\beta \in \mathcal{V}^*$

\noindent \textbf{Proof:} Since Step (2) of the method omits all axioms $\beta_i \in \mathcal{V}$ such that $\mathcal{O}, \beta_1,..., \beta_{i-1}, \beta_{i+1},..., \beta_n \models \beta_i$ via the annotation-based filtering followed by entailment checking on any remaining axioms, it follows that $\mathcal{O} \not\models \beta$ for every $\beta \in \mathcal{V}^*$. \\

Lemma 3 implies that $\mathcal{O} \not\models \mathcal{V}^*$.\\

\vspace{5pt}

\noindent \textbf{Lemma 4:} $\mathcal{O}, \mathcal{V}^* \models \mathcal{V} \setminus \mathcal{V}^*$

\noindent \textbf{Proof:} The extraction of $\mathcal{V}^*$ from $\mathcal{V}$ is performed sequentially. Thus, we can define a sequence: \\

\hfill $\mathcal{V}_0, \mathcal{V}_1, ... , \mathcal{V}_n$ \hfill \hphantom{(1)} \\

\noindent where $\mathcal{V}_0 = \mathcal{V}$, $\mathcal{V}_n = \mathcal{V}^*$ and for each $i$ with $0 \leq i < n$: \\

\hfill $\mathcal{O}, \mathcal{V}_i \setminus \beta_i \models \beta_i$ \hfill (1) \\

\noindent and \\

\hfill $\mathcal{V}_{i+1} = \mathcal{V}_i \setminus \beta_i$ \hfill (2) \\

\noindent where $\beta_i \in \mathcal{V} \setminus \mathcal{V}^*$ is the redundant axiom removed at step $i$. Now we can prove Lemma 4 by induction. Consider an axiom $\beta \in \mathcal{V} \setminus \mathcal{V}^*$, the base case is: \\

\hfill $\mathcal{O}, \mathcal{V}_0 \setminus \beta_0 \models \beta$ \hfill \hphantom{(1)} \\

\noindent There are two possible cases to consider. (i) $\beta_0 \neq \beta$, in which case $\beta \in \mathcal{V}_0 \setminus \beta_0$ and so the base case trivially holds. (ii) $\beta_0  = \beta$, in which case $\beta$ must be redundant according to the dual of Definition 1(iii) and thus statement (1) holds at step 0. We now define the induction hypothesis: \\

\hfill $\mathcal{O}, \mathcal{V}_i \setminus \beta_i \models \beta$ \hfill \hphantom{(1)} \\

\noindent and the induction step: \\

\hfill $\mathcal{O}, \mathcal{V}_{i+1} \setminus \beta_{i+1} \models \beta$ \hfill \hphantom{(1)} \\

There are three possible cases for the induction step. (i) $\beta \not\in \beta_0, ..., \beta_{i+1}$, i.e., the axiom $\beta$ has not yet been discarded as of step $i+1$. Then the induction step holds since $\beta \in \mathcal{V}_{i+1} \setminus \beta_{i+1}$. (ii) $\beta = \beta_{i+1}$, i.e., the axiom $\beta$ is removed at step $i+1$. Then statement (1) holds at step $i+1$ under the definition of redundancy in 1(iii), and thus the induction step holds. (iii) $\beta \in \beta_0, ..., \beta_i$, i.e., $\beta$ was checked and discarded prior to step $i+1$. Then from (1):\\

\hfill $\mathcal{O}, \mathcal{V}_{i+1} \setminus \beta_{i+1} \models \beta_{i+1}$ \hfill \hphantom{(1)} \\

\noindent and we can also write: \\

\hfill $\mathcal{O}, \mathcal{V}_{i+1} \setminus \beta_{i+1} \models \mathcal{O}, \mathcal{V}_{i+1} \setminus \beta_{i+1}, \beta_{i+1}$. \hfill \hphantom{(1)} \\

\noindent Which simplifies to $\mathcal{O}, \mathcal{V}_{i+1} \setminus \beta_{i+1} \models \mathcal{O}, \mathcal{V}_{i+1}$. By substituting statement (2) into this, we obtain: \\

\hfill $\mathcal{O}, \mathcal{V}_{i+1} \setminus \beta_{i+1} \models \mathcal{O}, \mathcal{V}_i \setminus \beta_i$ \hfill \hphantom{(1)} \\

\noindent From the induction hypothesis, $\mathcal{O}, \mathcal{V}_i \setminus \beta_i \models \beta$. Thus, the following holds: \\

\hfill $\mathcal{O}, \mathcal{V}_{i+1} \setminus \beta_{i+1} \models \beta$ \hfill \hphantom{(1)} \\

\noindent meaning that the induction step holds for all $\beta \in \mathcal{V} \setminus \mathcal{V}^*$. As a result, we have that:\\

\hfill $\mathcal{O}, \mathcal{V}_i \setminus \beta_i \models \beta$ \hfill \hphantom{(1)} \\

\noindent for all $0 \leq i < n$ and finally, by setting $i = n-1$ we have: \\

\hfill $\mathcal{O}, \mathcal{V}_{n-1} \setminus \beta_{n-1} \models \beta$ \hfill \hphantom{(1)} \\

\noindent by substituting (2):\\

\hfill $\mathcal{O}, \mathcal{V}_n \setminus \beta_{n-1} \models \beta$ \hfill \hphantom{(1)} \\

\noindent for all $\beta \in \mathcal{V} \setminus \mathcal{V}^*$. Since $\mathcal{V}_n = \mathcal{V^*}$: \\

\hfill $\mathcal{O}, \mathcal{V}^* \models \mathcal{V} \setminus \mathcal{V}^*$ \hfill \hphantom{(1)}\\

\noindent as required.

\vspace{5pt}

\noindent\textnormal{\textbf{Lemma 5:}} For any $\mathcal{W}$ in the signature $\mathcal{S}_A$ such that $\mathcal{O}, \beta_1,...,\beta_{i-1}, \beta_{i+1},...,\beta_n \not\models \beta_i$ for every $\beta_i \in \mathcal{W}$ and $\mathcal{O}, \lnot\psi \models \mathcal{W}$, if $\mathcal{O}, \mathcal{W} \models \mathcal{O}, \mathcal{V}^*$, \mbox{then $\mathcal{O}, \mathcal{V}^* \equiv \mathcal{O}, \mathcal{W}$}.

\vspace{5pt}

\noindent\textnormal{\textbf{Proof:}} We have that $\mathcal{O}, \lnot\psi \models \mathcal{W}$. We also have that $\mathcal{V} \models \mathcal{W}$, since $\mathcal{V}$ is the strongest necessary condition of $\mathcal{O}, \lnot\psi$ in the signature $\mathcal{S}_A$. We can write $\mathcal{V}$ as: \\

\hfill $\mathcal{V} = (\mathcal{V} \setminus \mathcal{V}^*), \mathcal{V}^*$ \hfill \hphantom{(0)} \\

\noindent and thus:\\ 

\hfill $(\mathcal{V} \setminus \mathcal{V}^*), \mathcal{V}^* \models \mathcal{W}$ \hfill \hphantom{(1)}\\ 

\noindent We can then write:\\

\hfill $\mathcal{O}, (\mathcal{V} \setminus \mathcal{V}^*), \mathcal{V}^* \models \mathcal{O}, \mathcal{W}$ \hfill (1) \\ 

\noindent From (1) and Lemma 4, we then derive: \\

\hfill $\mathcal{O}, \mathcal{V}^* \models \mathcal{O}, \mathcal{W}$ \hfill \hphantom{(0)} \\

\noindent as required.


\vspace{5pt}

Using Lemmas 1--5, it is possible to prove the soundness of the method with respect to the abduction problem in Definition 1.\\

\noindent \textbf{Theorem \ref{SoundnessAndCompleteness}}.\textit{Let $\mathcal{O}$ be an $\mathcal{ALC}$ ontology, $\psi$ an observation as a set of ABox axioms, excluding role assertions, and $\mathcal{S}_A$ a set of abducible symbols such that it includes all role symbols in $\mathcal{O}, \psi$ and $\mathcal{S}_A \subseteq sig(O, \psi)$.}

\noindent \textbf{(i) Soundness:} \textit{The hypothesis $\mathcal{H}$ returned by the method is a disjunction of ABox axioms such that sig($\mathcal{H}) \subseteq \mathcal{S}_A$ and $\mathcal{H}$ satisfies Definition 1(i)--(iv).}

\vspace{5pt}

\noindent \textbf{Proof:} We obtain the hypothesis $\mathcal{H}$ by negating $\mathcal{V}^*$ under contrapositive reasoning, which is then added to $\mathcal{O}$. Thus, condition (i) follows from Lemma 3 since $\beta_i \equiv \lnot\alpha_i$ for all $\beta_i \in \mathcal{V}^*$ and thus $\mathcal{O}, \alpha_i \not\models \perp$ for every disjunct $\alpha_i \in \mathcal{H}$. Condition (ii) follows from Lemma 2: since $\mathcal{O}, \lnot\psi \models \mathcal{V}^*$, under contraposition $\mathcal{O, H} \models \psi$ where $\mathcal{H} \equiv \lnot\mathcal{V}^*$. Condition (iii) is guaranteed via the strict check performed in Step (2) of the method, which is the dual of condition (iii). Thus, since $\mathcal{H}$ is obtained by applying contraposition to $\mathcal{V}^*$, and all axioms in $\mathcal{V}^*$ satisfy the check in Step (2), $\mathcal{H}$ will satisfy condition (iii). Condition (iv) follows from Lemma 5, which shows that if there exists a set of axioms $\mathcal{W}$ in the signature $\mathcal{S}_A$ such that $\mathcal{O}, \lnot\psi \models \mathcal{W}$, $\mathcal{W}$ satisfies the dual of condition (iii) and $\mathcal{O, W} \models \mathcal{O, V}^*$ then $\mathcal{O}, \mathcal{V}^* \equiv \mathcal{O, W}$. Since the hypothesis $\mathcal{H}$ is obtained by negating $\mathcal{V}^*$, the dual of Lemma 5 holds for $\mathcal{H}$: i.e., if there exists a $\mathcal{H}'$ such that $\mathcal{H}' = \lnot\mathcal{W}$ then $\mathcal{O, H} \equiv \mathcal{O, H}'$. \\


\noindent\textbf{(ii) Completeness:} \textit{If there exists a hypothesis $\mathcal{H}'$ such that $sig(\mathcal{H}') \subseteq \mathcal{S}_A$ and $\mathcal{H}'$ satisfies Definition 1(i)--(iv), then the method returns a hypothesis $\mathcal{H}$ such that $\mathcal{O, H} \equiv \mathcal{O, H}'$}.

\vspace{5pt}

\noindent \textnormal{\textbf{Proof:} This property follows directly from the interpolation completeness of $Int_{\mathcal{ALC}}$, see Section 4 (``Forgetting and Uniform Interpolation"). For any given combination of an ontology $\mathcal{O}$, negated observation $\lnot\psi$ and forgetting signature $\mathcal{F}$, a uniform interpolant $\mathcal{V}$ is returned using $Int_{\mathcal{ALC}}$ such that for any other uniform interpolant $\mathcal{V}'$ of  $(\mathcal{O}, \lnot \psi)^{-\mathcal{F}}$, the property  $\mathcal{V} \equiv \mathcal{V}'$ holds. Thus, the set $\mathcal{V}^*$ which satisfies the properties in Lemma 5 is always obtained from $\mathcal{V}$ as required. The result of applying contrapositive reasoning to $\mathcal{V}^*$ is then the hypothesis $\mathcal{H}$ such that for any other consistent hypothesis $\mathcal{H}'$ in the restricted signature $\mathcal{S}_A$ such that $\mathcal{O, H} \models \mathcal{O, H}'$, $\mathcal{O, H} \equiv \mathcal{O, H}'$.}

\subsubsection{Soundness of Annotation-based Filtering}
The annotation-based filtering method is used as preprocessing to reduce the cost of Step (2) by first computing $\mathcal{V}^*_{app}$ before applying the entailment check. Therefore, it is necessary to prove the soundness of this annotation-based filtering so that the soundness of the full method in practice is still guaranteed.

To do this, it is necessary to prove that the annotator concept $\mathcal{\ell}$ carries from the premises of an inference in $Int_{\mathcal{ALC}}$ to the conclusion, and thus that $\mathcal{\ell}$ will appear in any axiom in $\mathcal{V}$ that is dependent on $\lnot\psi$ in the presented method. By doing so, the soundness of the annotation-based filtering proposed in this paper is proved since this filtering method eliminates only those axioms in $\mathcal{V}$ that do not contain the concept $\mathcal{\ell}$.

To recall the notion of dependency: each axiom in the premise of a rule application in $Int_{\mathcal{ALC}}$ is referred to as a \emph{parent} of the conclusion of the rule. The \emph{ancestor} relation is then defined as the reflexive, transitive closure of the parent relation. Now dependency can be defined. An axiom is defined as \emph{dependent} on the negated observation $\lnot\psi$ if it has at least one ancestor axiom in $\lnot\psi$. 

\noindent \textbf{Theorem \ref{AnnotationCarries}:}
\textit{Let $\mathcal{O}$ be an ontology, $\psi$ an observation as a set of axioms, $\mathcal{F}$ a forgetting signature and $\mathcal{\ell}$ an annotator concept added as an extra disjunct to each axiom in $\lnot\psi$ where $\mathcal{\ell} \not\in sig(\mathcal{O, \psi})$ and $\mathcal{\ell} \not\in \mathcal{F}$. For every axiom  $\beta$ in the uniform interpolant $\mathcal{V} = (\mathcal{O}, \lnot\psi)^{-\mathcal{F}}$, $\beta$ is dependent on $\lnot\psi$ iff $\mathcal{\ell} \in sig(\beta)$.}

\vspace{5pt}

\noindent\textbf{Proof:} The proof is by induction over the way the derivation is constructed in $Int_{\mathcal{ALC}}$. 

The base case is the start of the derivation, where no inference has been performed yet. So we consider any clause $\beta$ in $(\mathcal{O}, \lnot\psi)$: the input to the abduction method. For an axiom $\beta$ to be dependent on $\lnot\psi$, it must have at least one ancestor in $Cls(\lnot\psi)$, where $Cls$ denotes the clausal form of $\lnot\psi$. Since in the base case no inferences have been performed, the only way for an axiom $\beta$ to have an ancestor in $Cls(\lnot\psi)$ is if $\beta \in Cls(\lnot\psi)$, as no other dependent axioms have been derived. Thus, the only axioms dependent on $\lnot\psi$ are those in the negated observation due to the reflexivity of the ancestor relation. 

Now consider the following set of axioms $\{\nu_1, ... ,\nu_k,\nu_{k+1}\}$ where each $\nu_i$ is the conclusion of an inference rule in $Int_{\mathcal{ALC}}$ between $\nu_{i-1}$ and another axiom, where $\nu_1 \in \lnot\psi$. Since we must prove a characteristic of dependent axioms, the inferences must all have at least one ancestor in $\lnot\psi$. Thus, the set of inferences begins with $\nu_1 \in \lnot\psi$. The induction hypothesis is that $\nu_k$ contains the annotator concept $\mathcal{\ell}$: i.e., $\mathcal{\ell} \in sig(\nu_k)$. Now for the induction step: two cases must be considered for the axiom $\nu_{k+1}$ . Given that one of the parents of $\nu_{k+1}$ is $\nu_k$, the other \mbox{parent $\beta$} can be (a) an axiom not dependent on $\lnot\psi$ or (b) another axiom that is dependent on $\lnot\psi$. In both cases, the inference can be made using any of the rules in Figure 1. The case where both parent axioms are not dependent on $\lnot\psi$ need not be considered, since the aim is to show that $\mathcal{\ell}$ is present in all axioms dependent on $\lnot\psi$ and the definition of dependency requires at least one ancestor to be in $\lnot\psi$.}\\

\noindent For case (a), where $\beta$ does not depend on $\lnot\psi$:

\begin{enumerate}[leftmargin=2em,label=\textbf{(\arabic*)}]
\item \textbf{Resolution:} Consider $\beta = (C_1 \lor \lnot C_2)(t_1)$ and $\nu_k = C_2 \lor \mathcal{\ell}(t_2)$ where $\sigma$ is the unifier of $t_1$ and $t_2$ if it exists. Resolution occurs on $C_2$ as follows:

\begin{figure}[h!]
	\parbox{\linewidth-1.1em}{
		{\centering{
		
		\underline{$(C_1 \lor \lnot C_2)(t_1) \hspace{20pt}  C_2 \lor \mathcal{\ell}(t_2)$}
		
		$(C_1 \lor \mathcal{\ell})\sigma$ 
		\vspace{1em}		
		
		therefore $\mathcal{\ell} \in sig(\nu_{k+1})$.
		
		}}

	}
\end{figure}

\vspace{-1em}

\item \textbf{Role propagation:} Two cases are considered: (i) $\nu_k$ contains an existential quantifier and (ii) $\nu_k$ contains a universal quantifier. Consider $\beta = (C \lor \forall r.D_1)(t_1)$ and (i) $\nu_k = \exists r.D_2 \lor \mathcal{\ell}(t_2)$, (ii) $\nu_k = \forall r.{D}_2 \lor \mathcal{\ell}(t_2)$ where $D_1$ and $D_2$ are definer symbols and $\sigma$ is the unifier of $t_1$ and $t_2$ if it exists. Role propagation occurs on $r$ as follows:

\begin{figure}[h!]
	\parbox{\linewidth-1.1em}{
		{\centering{
		(i)\hspace{0.75em}
		\underline{$(C \lor \forall r.D_1)(t_1) \hspace{20pt} \exists r.D_2 \lor \mathcal{\ell}(t_2)$}
		
		\hspace{0.75em}$(C \lor \mathcal{\ell})\sigma \lor \exists r.D_{12}(t_1\sigma)$ 
		
		\vspace{1em}
		
		therefore $\mathcal{\ell} \in sig(\nu_{k+1})$.
		
		}}
	
		\vspace{1em}
		{\centering{
		(ii)\hspace{0.75em}\underline{$(C \lor \forall r.D_1)(t_1) \hspace{20pt}  \forall r.{D}_2 \lor \mathcal{\ell}(t_2)$}
		
		\hspace{0.75em}$(C \lor \mathcal{\ell})\sigma \lor \forall r.D_{12}(t_1\sigma)$ 
	    \vspace{1em}		
		
		therefore $\mathcal{\ell} \in sig(\nu_{k+1})$.
		
		}}

}
\end{figure}

\item \textbf{Existential role restriction elimination:} Only one case needs to be considered for $\nu_k$, which is: $\nu_k = (C \lor \mathcal{\ell} \lor \exists r.D_1)(t)$. This is due to the fact that the annotator concept $\mathcal{\ell}$ is appended disjunctively to clauses in $\lnot\psi$ and thus does not occur under quantifiers. Thus, no definer symbols will be introduced in place of concepts containing $\mathcal{\ell}$. Let $\beta = \lnot D(x)$, then existential role restriction elimination is applied as follows:

\begin{figure}[h!]
	\parbox{\linewidth-1.1em}{
		{\centering{
		
		\underline{$(C \lor \mathcal{\ell} \lor \exists r.D_1)(t) \hspace{20pt}  \lnot D_1(x)$}
		
		$C \lor \mathcal{\ell}$ 
		\vspace{1em}
		
		therefore $\mathcal{\ell} \in sig(\nu_{k+1})$.		
		
		}}
				
	}
\end{figure}

\item \textbf{Role instantiation:} Since the observation may not contain role assertions, only one case needs to be considered for $\nu_k$, which is: $\nu_k = (C_1 \lor \ell \lor (\forall .D)(t_1)$. Let $\beta = r(t_2, b)$, then role instantiation is applied as follows:

\begin{figure}[h!]
	\parbox{\linewidth-1.1em}{
		{\centering{
		
		\underline{$(C_1 \lor \ell \lor (\forall r.D))(t_1)$ \hspace{20pt}  $r(t_2, b)$}
		
		$(C_1 \lor \ell )\sigma \lor D(b)$ 
		\vspace{1em}
		
		therefore $\mathcal{\ell} \in sig(\nu_{k+1})$.		
		
		}}
				
	}
\end{figure}

\end{enumerate}

\vspace{-1em}

\noindent For case (b), where $\beta$ is dependent on $\lnot\psi$: the derivations are largely the same. For the resolution and role propagation rule the result of the inference simply contains $\mathcal{\ell} \lor \mathcal{\ell}$, which simplifies to $\mathcal{\ell}$. For the existential role restriction elimination and role instantiation rules, it is not possible for the second axiom to also be dependent on $\lnot\psi$. This is due to the fact that $\mathcal{\ell}$ does not fall under the scope of a quantifier, thus there will be no clause of the form $\lnot D \lor \mathcal{\ell}$, where $D$ is a definer, introduced during the transformation to the normal form required by $Int_{\mathcal{ALC}}$. Additionally, $\psi$ does not contain negated role assertions, so there will also be no clause of the form $r(a, b) \lor \mathcal{\ell}$. \\

\noindent Thus, each axiom in $\mathcal{V}$ that has at least one ancestor in $\lnot\psi$ will contain the annotator concept $\mathcal{\ell}$. Since $\mathcal{\ell}$ is not present in $\mathcal{O}$, having been disjunctively appended to $\lnot\psi$, the approximation of the reduced uniform interpolant $\mathcal{V}^*$ obtained by eliminating all axioms $\beta_i$ such that $\mathcal{\ell} \not\in sig(\beta_i)$ will always take the form of a set of axioms that are dependent on $\lnot\psi$.

\subsection{Complexity}
The main source of complexity for the presented abduction method is in the use of the forgetting method $Int_{\mathcal{ALC}}$. Computing the uniform interpolant $\mathcal{V}$ has 2EXPTIME complexity and the number of clauses in the uniform interpolant is double exponential in the size of the input ontology \cite{KoopmannABoxes}.

As for the extraction of hypotheses from uniform interpolants, this is done by first approximating $\mathcal{V}^*$ using the annotation-based filtering method described in the previous section, resulting in $\mathcal{V}^*_{app}$. This relies on checking the signature of each axiom in $\mathcal{V}$ for the presence of the annotator concept $\mathcal{\ell}$. The complexity of this filtering method is linear in the size of the uniform interpolant $\mathcal{V}$. 

In the case where the fully reduced uniform interpolant $\mathcal{V}^*$ is computed, the additional check $\mathcal{O}, \{\beta_1,..., \beta_{i-1}, \beta_{i+1},..., \beta_n\} \not\models \beta_i$ for each remaining axiom $\beta_i$ in $\mathcal{V}^*_{app}$ is needed to remove remaining redundancies. In the worst case, $\mathcal{V}^*_{app}$ could be equal to $\mathcal{V}$, and thus could be double exponential in size with respect to $(\mathcal{O}, \psi)$. In this case, this additional step would require a double exponential number of exponential time entailment checks. Thus, the worst case complexity of this step is 3EXPTIME. In practice, however, the size of $\mathcal{V}^*_{app}$ is much smaller than that of $\mathcal{V}$, as shown by the experiments in Table 2.

Thus, the overall worst case time complexity of the proposed abduction method is 3EXPTIME, and the size of the hypothesis produced is at most double exponential with respect to the input ontology, which in this case is $\mathcal{O}, \lnot\psi$ where $\lnot\psi$ is the negation of the observation.



\end{document}